\documentclass[11pt]{article}
\usepackage{arxiv}
\usepackage{booktabs}
\usepackage{graphicx}
\usepackage{amsmath}
\usepackage{amssymb}
\usepackage{hyperref}

\title{A Cross-Model VLM-Judge Protocol for Single-Image 3D Mesh Quality
(and Why Cheap Proxies Fall Short)}

\author{%
  Ali Asaria \\ Transformer Lab \and
  Tony Salomone \\ Transformer Lab \and
  Deep Gandhi\thanks{Corresponding author: \texttt{deep@lab.cloud}} \\ Transformer Lab
}
\date{}
\runningtitle{Judging Single-Image 3D Generation}

\begin{document}
\maketitle

\begin{abstract}
Single-image-to-3D generators are improving quickly, but there is no agreed,
human-free way to tell whether one generated mesh is better than another.
Practitioners commonly rely on cheap automatic proxies (render-space CLIP
similarity and mesh geometry-validity statistics), yet how well these track
perceived quality is unestablished. We make two contributions. First, we propose
and validate a reproducible \emph{VLM-judge evaluation protocol}: a fixed
24-view headless render rig, \emph{two independent} vision-language judge
families, and a mandatory position-bias correction that queries both
presentation orders and keeps only order-consistent verdicts. The two judge families agree \emph{substantially} with each other (Cohen's
$\kappa{=}0.66$), well above the chance-agreement floor. Second, using this protocol as the reference, we show the cheap proxies do
\emph{not} substitute for it. Geometry validity is only a \emph{weak} signal on
average (because, as we show, it is bimodal) and stays below our pre-registered
target, while render-CLIP is \emph{at chance}. A learned Bradley--Terry head collapses onto a single manifoldness statistic
(giving render-CLIP a negative weight) and matches geometry-only exactly, so
learning the feature weights buys nothing. The proxy is also \emph{bimodal}: it is significantly above chance on contrasts
with visible geometric defects but at chance on ambiguous contrasts, consistent with geometry validity tracking the judge only when the defect is
visually salient. We therefore recommend the VLM-judge protocol as a reliable, reproducible
evaluator \emph{under the conditions tested} (two feed-forward generators on
Google Scanned Objects, with a face-drop degradation regime) and advise against
geometry/CLIP proxies as optimization targets.
\end{abstract}

\section{Introduction}
Single-image-to-3D generators now produce plausible textured meshes from one
photo, but evaluating them remains ad hoc. Cherry-picked demos look strong;
quantitative comparison defaults to render-space CLIP similarity or to mesh
geometry-validity statistics (watertightness, manifoldness, normal
consistency), because both are cheap and need no human labels. Whether these
proxies actually track \emph{perceived} mesh quality, the thing a downstream
user, ranker, or training signal cares about, has not been carefully measured.

We take the position that the right reference for ``is this mesh good?'' is a
\emph{vision-language-model judge} applied to a fixed multi-view render of the
asset, and that the central engineering questions are how to make such a judge
\emph{reliable} and how to test whether cheaper proxies can stand in for it. We
build a protocol around two independent open VLM judge families (an oracle and a
separate validation judge)
on a fixed 24-view headless render rig, with a position-bias correction that turns out to be essential. We then use the
validated judge as the reference against which the cheap proxies (geometry
validity, render-CLIP, their composite, and a learned combination) are scored
on strictly held-out objects.

Our finding is that the protocol is reliable (substantial cross-model agreement)
while the proxies are weak, and that the proxies fail in a specifically
misleading way: they are significantly above chance only on contrasts where the
geometric defect is \emph{visible} in the render (telling two generators apart,
or a clearly-holed mesh from an intact one) and fall to chance on the more
ambiguous quality calls that matter for ranking or optimizing a single model.

\paragraph{Contributions.}
\begin{enumerate}
\item \textbf{A validated VLM-judge evaluation protocol} for single-image-to-3D
mesh quality (fixed render rig, two independent judge families, and a
swap-and-keep-consistent position-bias correction), whose judge families agree at
$0.83$ (Cohen's $\kappa{=}0.66$, substantial; chance floor $0.51$), well above a
trivial-agreement baseline (Table~\ref{tab:headline}).
\item \textbf{Evidence that position bias is large and must be corrected}: about
$26\%$ of raw verdicts flip with presentation order (and this fraction is itself
sample-dependent); as one illustrative baseline, correcting it moved an agreement
estimate from $0.333$ to $0.714$ (\S\ref{sec:method}).
\item \textbf{A demonstration that cheap proxies do not substitute for the
judge}: geometry validity is significantly above chance but \emph{weak} ($0.62$,
$[0.55,0.69]$) and below our target; render-CLIP is at chance ($0.48$); and a
learned head collapses onto a single manifoldness statistic, so learning the
weights buys nothing (Table~\ref{tab:headline}).
\item \textbf{A subgroup/error analysis} showing the proxy is bimodal: it is
significantly above chance on visible-defect contrasts (cross-generator $0.91$,
within-TripoSR $0.80$) but at chance on ambiguous ones (cross-generator-mixed
$0.53$; $z{=}5.06$ for the gap), which we interpret as geometry validity tracking
the judge only when the defect is visually salient (\S\ref{sec:results},
\S\ref{sec:discussion}).
\end{enumerate}

\section{Related Work}
\paragraph{Single-image and multi-view 3D generation.}
Fast feed-forward single-image-to-3D pipelines such as Unique3D~\citep{2405.20343v3}
and asset pipelines like Meta~3D~Gen~\citep{2407.02599v1} have made per-object
generation cheap enough to evaluate at scale, and studies of 3D
representations~\citep{2509.02474v1} and photorealistic
generation~\citep{2605.13852v1} highlight how differently generators behave. We
treat such generators as black boxes and focus on \emph{how to judge} their
outputs.

\paragraph{Evaluating 3D generation.}
Benchmarks for multi-view generation~\citep{2507.00006v1} and rethinking of
point-cloud generation metrics~\citep{2511.05308v2} both note that standard
automatic metrics correlate weakly with perceived quality; curated-quality
datasets~\citep{2504.07334v2} and large object corpora~\citep{2603.16866v1}
provide inputs but not a quality oracle. Empirical studies of memorization in 3D
shape generation~\citep{2512.23628v1} and robust shape
generation~\citep{2601.11514v1} further motivate evaluation that looks at the
rendered asset rather than at a single scalar. Our protocol contributes a
reusable, human-free reference and a direct measurement of how far the cheap
proxies fall short of it.

\paragraph{VLM-as-judge.}
Using vision-language models as judges is now common for image
generation~\citep{2603.27862v1}, but their reliability is contested:
\citet{2603.07990v2} report that even frontier VLMs are imperfect judges on
grounded multimodal tasks, and presentation-order (position) bias is a known
failure mode of LLM/VLM judges~\citep{2306.05685,2305.17926}. Our contribution is not these techniques in the
abstract but a \emph{validated} instantiation for the 3D-render setting:
cross-model agreement between \emph{two different} judge families together with
swap-consistency position-bias correction, quantified by Cohen's $\kappa$ against
a chance-agreement baseline, and a measurement of how far cheap proxies fall short
of it.

\paragraph{Aligning 3D generators with rewards.}
A growing line uses reward or preference signals to align 3D
generators: simulation feedback for physical soundness~\citep{2503.22677v2},
2D-reward diffusion alignment~\citep{2506.15684v1}, and direct preference
optimization from human preferences~\citep{2502.04370v1}, with related
preference-optimization~\citep{2511.20629v5} and parameter-efficient
adaptation~\citep{2504.15933v2} machinery. Our results bear directly on this
line: we find that the cheap automatic proxies one might use as such a reward are
a weak quality signal (geometry significantly above chance but well below target;
render-CLIP at chance), so reward-based specialization should be driven by the
(de-biased) VLM-judge preferences themselves rather than by a proxy. We treat
generator specialization as out of scope and future work.

\section{Method}
\label{sec:method}
\paragraph{The evaluation protocol.}
Given a generated mesh, we normalize it to a unit bounding box and render a fixed
$24$-view turntable rig with a headless offscreen rasterizer. Quality is a \emph{pairwise} judgment between two meshes generated from the same
input image. We use \emph{two different} open VLM families: an oracle judge
$X$ (Qwen2.5-VL-7B-Instruct~\citep{2502.13923}) and an independent validation
judge $Y$ (InternVL3-8B~\citep{2504.10479}). Keeping $X\neq Y$ lets us report cross-model agreement as a reliability check
rather than trusting a single model.

\paragraph{Position-bias correction.}
VLM judges exhibit a presentation-order bias. For each pair we query the judge in
\emph{both} orders ($A,B$ and $B,A$) and keep the verdict only if it is consistent
across the swap; order-dependent verdicts are discarded as position-biased. This is not optional: about $26\%$ of raw verdicts are inconsistent (a fraction
that is itself sample-dependent, $0.58$--$0.63$ consistent at smaller $N$); as one illustrative baseline, the uncorrected agreement reads $0.333$ versus
$0.714$ after correction.

\paragraph{Proxy rewards under test.}
We score the same renders/meshes with the cheap proxies a practitioner might use.
For a mesh we extract five features: watertightness, manifoldness,
non-self-intersection, normal consistency, and render-CLIP similarity
(CLIP~\citep{2103.00020}, open\_clip ViT-B-32~\citep{ilharco_openclip}). From these we form: (i) a geometry-only score, (ii) a render-CLIP-only score, (iii) a
fixed-weight composite, and (iv) a \emph{learned} pairwise Bradley--Terry
head~\citep{bradley1952} fit on judge-$X$ labels:
\begin{equation}
P(a \succ b) = \sigma\!\big(w^\top (\phi(a) - \phi(b))\big),
\end{equation}
where $\phi(\cdot)$ is the five-feature vector and $w$ is learned by logistic
regression~\citep{scikit-learn}. All proxies are evaluated against the independent
judge $Y$.

\section{Experimental Setup}
\label{sec:setup}
\paragraph{Data and generators.}
Inputs are single-view photos from Google Scanned Objects~\citep{2204.11918}
(public, CC-BY 4.0; $3{,}069$ object directories available). Per object we form four candidates spanning a quality
spectrum: meshes from two single-image-to-3D generators run as black boxes (Stable
Fast~3D~\citep{2408.00653} and TripoSR~\citep{2403.02151}), plus a face-dropped degraded variant of each. We sample $N{=}60$ objects for the
main evaluation.

\paragraph{Splits and determinism.}
Splits are strictly by object so no object straddles train and test, and the
oracle judge $X$ is never the validation judge $Y$. The render rig and candidate construction are deterministic given a fixed
seed, and both judges use greedy (argmax) decoding. The full corpus is $262$
position-consistent pairs at $N{=}60$. The rule-based signals (geometry, render-CLIP, the fixed composite) are
\emph{not} fit to any data, so we evaluate them descriptively over the full
corpus; the strict by-object train/test split matters only for the \emph{learned}
head, which is trained on train-object pairs and compared on the held-out
test-object pairs ($98$ pairs) against judge $Y$.

\paragraph{Statistical analysis.}
We report proportions with Wilson 95\% confidence intervals~\citep{wilson1927} and test agreement
against the chance rate $0.5$ with a two-sided binomial test. Because the unit of
observation (a pair) is clustered within objects (four candidates, hence six
pairs, per object), the effective sample is the $60$ objects rather than the
$98$--$262$ pairs; our headline intervals therefore use a cluster
bootstrap~\citep{efron1979} that
resamples \emph{objects}. Judge--judge agreement is additionally summarized by Cohen's
$\kappa$~\citep{cohen1960} relative
to the marginal-agreement baseline. The cluster bootstrap is used for the
headline geometry and agreement intervals; the per-subgroup and CLIP CIs in
Tables~\ref{tab:headline}--\ref{tab:subgroup} are pair-level Wilson intervals,
which are anti-conservative under clustering and so should be read together with
the exploratory caveat below. The four subgroup cells
($n{=}43/50/119/50$) are \emph{exploratory}; we make a large family of
accuracy-vs-chance comparisons and do not apply a formal multiple-comparison
correction, so single-cell directional readings should be treated as
hypothesis-generating.

\paragraph{Compute.}
The study used $\approx 3.4$ H100-hours across $14$ completed jobs (one further
run discarded) on a single H100-class GPU (predominantly H100-SXM5).

\section{Results}
\label{sec:results}

\begin{figure}[t]
\centering
\includegraphics[width=0.85\linewidth]{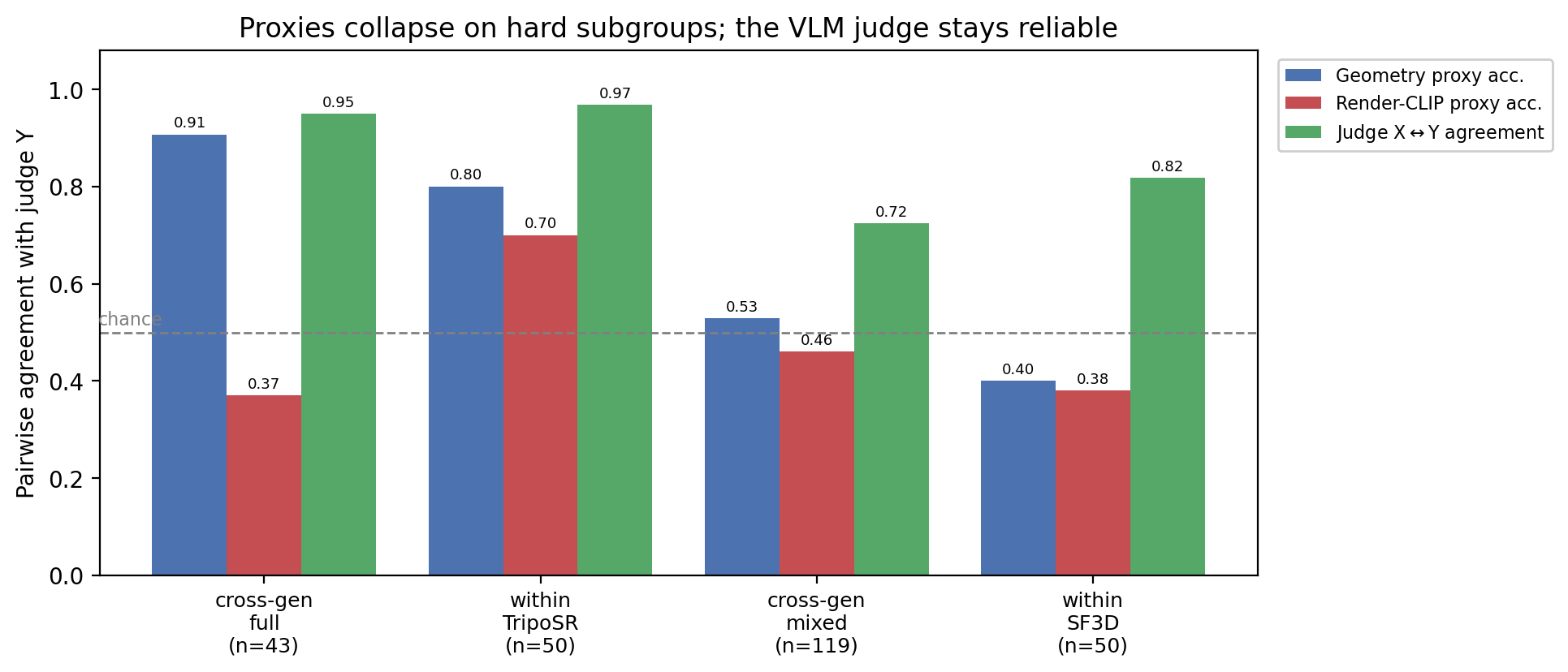}
\caption{Cheap proxies collapse on the ambiguous subgroups while the VLM judge
stays reliable. Geometry-proxy accuracy (blue) is significantly above chance on
the visible-defect contrasts (cross-generator-full, within-TripoSR) but at chance
on cross-generator-mixed (within-SF3D is confounded; see text); render-CLIP (red)
is weak throughout; cross-model judge agreement (green) stays $0.72$--$0.97$.
Dashed line is chance ($0.5$); error bars / small $n$ mean single-cell readings
are exploratory.}
\label{fig:subgroup}
\end{figure}

\paragraph{The judge protocol is reliable.}
The two independent judge families agree with each other on $0.83$ of $120$
dual-labeled pairs (Wilson 95\% CI $[0.76,0.89]$). Because a forced-choice verdict has a $0.5$ chance rate (and the marginal
agreement floor here is $0.51$), we summarize agreement by Cohen's
$\kappa{=}0.66$, ``substantial'' on the usual scale~\citep{landis1977}, confirming the agreement is
not a trivial artifact of skewed marginals. Agreement is above chance in every subgroup, and higher on contrasts with a
clear quality difference than on ambiguous ones ($0.72$ on cross-generator-mixed
vs.\ $0.95$--$0.97$ on the clearest cells; the $0.72$ cell is above chance but
not, on its own $n$, distinguishable from the others)
(Table~\ref{tab:subgroup}, Fig.~\ref{fig:subgroup}). We did not run the optional human spot-check, so reliability here means
cross-model consistency, not validated agreement with human raters.

\paragraph{The proxies do not substitute for the judge.}
Geometry validity is a \emph{weak but real} signal: across the full corpus it
agrees with the judge on $0.62$ of pairs (Wilson $[0.56,0.68]$; cluster-bootstrap
over objects $[0.55,0.69]$; two-sided binomial $p{<}0.001$ vs.\ $0.5$), so it is
significantly above chance yet far below our pre-registered target of
$\approx{\ge}0.75$. Render-CLIP, by contrast, is \emph{at chance} ($0.48$, $[0.42,0.54]$,
$p{=}0.50$); we therefore cannot claim it is anti-correlated, only that it
carries no usable quality signal. The learned Bradley--Terry head does not help: it places almost all weight on
manifoldness ($2.16$) and a \emph{negative} weight on render-CLIP ($-0.11$), and
the resulting ranking matches geometry-only exactly (zero lift at both $N{=}30$
and $N{=}60$): given freedom to weight the features, it collapses onto a single geometry
statistic. We read this as evidence that the limit lies in this feature set rather
than in the model, though we cannot rule out that richer features would do better.
On the strict held-out test subset ($98$ pairs) the same three rewards drop to
$0.52$ and render-CLIP to $0.40$; this subset is smaller and noisier than the full corpus, and the downward shift
from larger in-sample estimates ($0.66$ at $N{=}24$) is consistent with
small-sample optimism rather than a demonstrated trend.

\begin{table}[t]
\centering
\caption{Agreement with the independent validation judge $Y$, with Wilson 95\%
CIs. Geometry validity is significantly above chance but weak; render-CLIP is at
chance. The geometry-only, learned, and composite rewards produce effectively the
\emph{same} ranking (the learned head collapses onto manifoldness), so they are
not three independent corroborations. ``Full corpus'' = all $262$
position-consistent pairs (the rule-based signals are untrained); ``held-out'' =
the $98$ test pairs used for the learned head. Target is our pre-registered
directional goal, not a pass/fail gate.}
\label{tab:headline}
\begin{tabular}{lrrr}
\toprule
Reward & Full corpus & Held-out & vs.\ target \\
\midrule
Geometry-only        & $0.62\ [0.56,0.68]$ & $0.52$ & $\approx{\ge}0.75$ \\
Learned head (BT)    & ($\equiv$ geometry) & $0.52$ & $\approx{\ge}0.75$ \\
Fixed composite      & ($\equiv$ geometry) & $0.52$ & $\approx{\ge}0.75$ \\
Render-CLIP only     & $0.48\ [0.42,0.54]$ & $0.40$ & $\approx{\ge}0.75$ \\
\midrule
Judge $X\leftrightarrow Y$ agreement & \multicolumn{2}{c}{$0.83\ [0.76,0.89]$, $\kappa{=}0.66$} & n/a \\
\bottomrule
\end{tabular}
\end{table}

\paragraph{Why this is dangerous: a bimodal subgroup pattern.}
Table~\ref{tab:subgroup} and Fig.~\ref{fig:subgroup} break accuracy down by pair
type. Geometry validity is significantly above chance on the two contrasts with
\emph{visible} geometric defects: cross-generator (Stable Fast~3D vs.\ TripoSR,
$0.91$, $[0.78,0.96]$, $p{<}0.001$) and within-TripoSR degradation ($0.80$,
$[0.67,0.89]$, $p{<}0.001$). It is only at chance on cross-generator-mixed pairs
($0.53$, $[0.44,0.62]$, $p{=}0.58$). The cross-generator-vs-ambiguous gap is large and significant (two-proportion
$z{=}5.06$, $p{<}0.0001$). The within-Stable-Fast-3D cell ($0.40$, $[0.28,0.54]$, $p{=}0.20$) is
\emph{not} statistically below chance and is confounded (see Limitations), so we
do not lean on it. Render-CLIP is weak throughout
($0.37$--$0.70$). The danger is that the easy, visible-defect regime (exactly the setting in
which proxies are usually reported to ``work'') hides the chance-level behavior
on the more ambiguous calls that matter for ranking or optimizing a single
model.

\begin{table}[t]
\centering
\caption{Per-subgroup geometry-proxy agreement with judge $Y$ (Wilson 95\% CIs).
Geometry is significantly above chance only on the visible-defect contrasts
(cross-generator-full, within-TripoSR); on cross-generator-mixed it is at chance,
and the within-SF3D cell is not significantly below chance and is confounded.
Cells are exploratory ($n$ shown). Render-CLIP and judge agreement given for
reference.}
\label{tab:subgroup}
\begin{tabular}{lrrrr}
\toprule
Subgroup & $n$ & Geometry [95\% CI] & R-CLIP & Judge $X{\leftrightarrow}Y$ \\
\midrule
Cross-generator, full   & $43$  & $0.91\ [0.78,0.96]$ & $0.37$ & $0.95$ \\
Within-TripoSR          & $50$  & $0.80\ [0.67,0.89]$ & $0.70$ & $0.97$ \\
Cross-generator, mixed  & $119$ & $0.53\ [0.44,0.62]$ & $0.46$ & $0.72$ \\
Within-Stable Fast 3D$^\dagger$ & $50$ & $0.40\ [0.28,0.54]$ & $0.38$ & $0.82$ \\
\bottomrule
\end{tabular}
\\[2pt]
{\footnotesize $^\dagger$ confounded: the judge's own intact-vs-degraded call is
itself at chance here ($0.40\,[0.28,0.54]$), so this cell measures reference
noise as much as proxy failure.}
\end{table}

\paragraph{Error mechanism (interpretation).}
The subgroup pattern is \emph{consistent with} a visual-salience account. On
within-Stable-Fast-3D pairs the judge prefers the intact mesh only $0.40$ of the
time ($[0.28,0.54]$, i.e.\ itself at chance), versus $0.80$ for within-TripoSR,
and prefers Stable Fast~3D over TripoSR only $0.09$ of the time. Stable Fast~3D meshes are open shells, so dropping faces is plausibly not
clearly visible in a rendered view, whereas TripoSR meshes are watertight and
degradation shows up as holes. Geometry validity always penalizes the degraded
mesh, but the judge appears to agree mainly when the geometric defect is
\emph{visually salient}. We did not run the controlled manipulation (e.g.\
re-rendering at higher resolution or measuring silhouette difference) that would
turn this from an observational correlation into a tested mechanism, so we offer
it as our interpretation rather than a demonstrated cause; either way, the
within-SF3D cell is confounded by stimulus visibility and is excluded from our
proxy-failure claim.

\begin{figure}[t]
\centering
\includegraphics[width=0.5\linewidth]{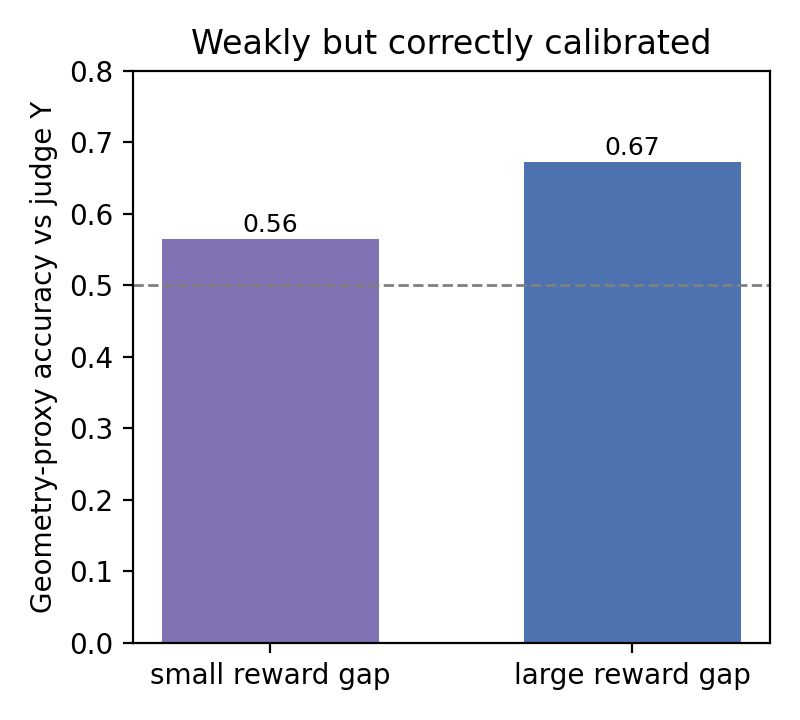}
\caption{The geometry proxy is weakly but correctly calibrated: accuracy is
higher when the reward gap between candidates is large. Dashed line is chance.}
\label{fig:calibration}
\end{figure}

\paragraph{Calibration.}
There is a hint that the geometry proxy is more often right on large-reward-gap
pairs ($0.67$) than small-gap pairs ($0.57$) (Fig.~\ref{fig:calibration}), but with $n{\approx}49$ per bin this difference is not statistically
significant (overlapping CIs); we report it as directionally suggestive only.

\section{Discussion and Limitations}
\label{sec:discussion}
The positive result is a usable, reproducible evaluator: a fixed render rig, two
independent VLM judge families, and a position-bias correction together give
cross-model agreement of $0.83$ ($\kappa{=}0.66$, substantial), with no human labels. The cautionary result is that the cheap proxies one might
reach for instead are a weak signal at best ($0.62$ for geometry, render-CLIP at
chance) and, more subtly, look deceptively good on the visible-defect contrasts
that are most often reported while dropping to chance on the ambiguous
ones. A practical recommendation follows for the reward-alignment
literature~\citep{2503.22677v2,2506.15684v1,2502.04370v1}, \emph{within the regime
we tested} (two feed-forward generators on Google Scanned Objects with a
face-drop degradation): drive specialization with the de-biased VLM-judge
preferences directly rather than with a geometry/CLIP proxy reward, which a
generator could satisfy (e.g.\ by maximizing manifoldness) without improving
perceived quality.

\subsection{Limitations}
\paragraph{The judge is the oracle.} We treat the VLM judge as ground truth, so
``quality'' is defined relative to these judges. We mitigate by requiring
agreement between two independent families, correcting position bias, and
reporting agreement rather than assuming a gold label, but VLM judges remain
imperfect~\citep{2603.07990v2}.

\paragraph{One subgroup is confounded.} On within-SF3D pairs the judge's own
intact-vs-degraded call is itself at chance (face-dropping an open shell is barely
visible at $256$px), so we exclude that cell and rest the bimodality claim on the significant
cross-generator and within-TripoSR cells.

\paragraph{Narrow regime.} The face-drop degradation injects exactly the
missing-geometry defect geometry statistics detect and CLIP misses, so it may
flatter geometry; and we test only two feed-forward generators, one object source
(Google Scanned Objects), and $N{=}60$ objects. Naturalistic failures (thin structures, transparency, multi-object inputs) and
structurally different generators are future work; our intervals reflect
object-level clustering but not generator or dataset transfer.

\section{Availability}
The evaluation protocol and per-pair data are available from the authors on
request. No model checkpoints are released; the generators and judges used are
existing public models.

\section{Conclusion and Future Work}
For single-image-to-3D, a cross-model, position-bias-corrected VLM-judge protocol
is a reliable, reproducible human-free evaluator ($\kappa{=}0.66$) under the
conditions we tested, while cheap geometry/CLIP proxies are weak (geometry $0.62$,
below target; render-CLIP at chance) and misleading on the visible-defect
contrasts where they are usually reported. Future work: richer learned visual
representations over the render rig (rather than CLIP) as a candidate automatic
reward, broader generator/object coverage and naturalistic failure modes, a human
spot-check to validate the judge against people, and (in a separate effort)
generator specialization driven directly by the de-biased judge preferences.

\bibliographystyle{unsrtnat}
\bibliography{references}

\begin{thebibliography}{32}
\providecommand{\natexlab}[1]{#1}
\providecommand{\url}[1]{\texttt{#1}}
\expandafter\ifx\csname urlstyle\endcsname\relax
  \providecommand{\doi}[1]{doi: #1}\else
  \providecommand{\doi}{doi: \begingroup \urlstyle{rm}\Url}\fi

\bibitem[Wu et~al.(2024)Wu, Liu, Cai, Yan, Wang, Hu, Duan, and
  Ma]{2405.20343v3}
Kailu Wu, Fangfu Liu, Zhihan Cai, Runjie Yan, Hanyang Wang, Yating Hu, Yueqi
  Duan, and Kaisheng Ma.
\newblock {Unique3D}: High-quality and efficient 3d mesh generation from a
  single image.
\newblock 2024.
\newblock URL \url{https://arxiv.org/abs/2405.20343}.

\bibitem[Bensadoun et~al.(2024)Bensadoun, Monnier, Kleiman, Kokkinos, Siddiqui,
  Kariya, Harosh, Shapovalov, Graham, Garreau, Karnewar, Cao, Azuri, Makarov,
  Le, Toisoul, Novotny, Gafni, Neverova, and Vedaldi]{2407.02599v1}
Raphael Bensadoun, Tom Monnier, Yanir Kleiman, Filippos Kokkinos, Yawar
  Siddiqui, Mahendra Kariya, Omri Harosh, Roman Shapovalov, Benjamin Graham,
  Emilien Garreau, Animesh Karnewar, Ang Cao, Idan Azuri, Iurii Makarov,
  Eric-Tuan Le, Antoine Toisoul, David Novotny, Oran Gafni, Natalia Neverova,
  and Andrea Vedaldi.
\newblock {Meta 3D Gen}.
\newblock 2024.
\newblock URL \url{https://arxiv.org/abs/2407.02599}.

\bibitem[Wiedemann et~al.(2025)Wiedemann, Liu, Leboutet, Gao, Ummenhofer,
  Paulitsch, and Yuan]{2509.02474v1}
Nina Wiedemann, Sainan Liu, Quentin Leboutet, Katelyn Gao, Benjamin Ummenhofer,
  Michael Paulitsch, and Kai Yuan.
\newblock {Unifi3D}: A study on 3d representations for generation and
  reconstruction in a common framework.
\newblock 2025.
\newblock URL \url{https://arxiv.org/abs/2509.02474}.

\bibitem[Sobol et~al.(2026)Sobol, Sohn, Blum, Zakharov, Bluvstein, Vedaldi, and
  Litany]{2605.13852v1}
Ido Sobol, Kihyuk Sohn, Yoav Blum, Egor Zakharov, Max Bluvstein, Andrea
  Vedaldi, and Or~Litany.
\newblock {Realiz3D}: 3d generation made photorealistic via domain-aware
  learning.
\newblock 2026.
\newblock URL \url{https://arxiv.org/abs/2605.13852}.

\bibitem[Xie et~al.(2025)Xie, Zou, Karumuri, Lenssen, and
  Pons-Moll]{2507.00006v1}
Xianghui Xie, Chuhang Zou, Meher~Gitika Karumuri, Jan~Eric Lenssen, and Gerard
  Pons-Moll.
\newblock {MVGBench}: Comprehensive benchmark for multi-view generation models.
\newblock 2025.
\newblock URL \url{https://arxiv.org/abs/2507.00006}.

\bibitem[Bastico et~al.(2025)Bastico, Ryckelynck, Cort\'e, Tillier, and
  Decenci\`ere]{2511.05308v2}
Matteo Bastico, David Ryckelynck, Laurent Cort\'e, Yannick Tillier, and Etienne
  Decenci\`ere.
\newblock {Rethinking Metrics and Diffusion Architecture for 3D Point Cloud
  Generation}.
\newblock 2025.
\newblock URL \url{https://arxiv.org/abs/2511.05308}.

\bibitem[Lin et~al.(2025)Lin, Liu, Lin, Bright, Tang, He, Liu, Zhu, and
  Le]{2504.07334v2}
Chendi Lin, Heshan Liu, Qunshu Lin, Zachary Bright, Shitao Tang, Yihui He,
  Minghao Liu, Ling Zhu, and Cindy Le.
\newblock {Objaverse++}: Curated 3d object dataset with quality annotations.
\newblock 2025.
\newblock URL \url{https://arxiv.org/abs/2504.07334}.

\bibitem[Wang et~al.(2026)Wang, Chen, Liu, Su, Zhu, Wang, Li, Chen, ang Gao,
  Qin, Wang, Zhang, Xu, Yu, Mu, and Luo]{2603.16866v1}
Kaixuan Wang, Tianxing Chen, Jiawei Liu, Honghao Su, Shaolong Zhu, Minxuan
  Wang, Zixuan Li, Yue Chen, Huan ang Gao, Yusen Qin, Jiawei Wang, Qixuan
  Zhang, Lan Xu, Jingyi Yu, Yao Mu, and Ping Luo.
\newblock {ManiTwin}: Scaling data-generation-ready digital object dataset to
  100k.
\newblock 2026.
\newblock URL \url{https://arxiv.org/abs/2603.16866}.

\bibitem[Pu et~al.(2025)Pu, Zeng, Zhou, Wang, and Liu]{2512.23628v1}
Shu Pu, Boya Zeng, Kaichen Zhou, Mengyu Wang, and Zhuang Liu.
\newblock {Memorization in 3D Shape Generation: An Empirical Study}.
\newblock 2025.
\newblock URL \url{https://arxiv.org/abs/2512.23628}.

\bibitem[Siddiqui et~al.(2026)Siddiqui, Frost, Aroudj, Avetisyan,
  Howard-Jenkins, DeTone, Moulon, Wu, Li, Straub, Newcombe, and
  Engel]{2601.11514v1}
Yawar Siddiqui, Duncan Frost, Samir Aroudj, Armen Avetisyan, Henry
  Howard-Jenkins, Daniel DeTone, Pierre Moulon, Qirui Wu, Zhengqin Li, Julian
  Straub, Richard Newcombe, and Jakob Engel.
\newblock {ShapeR}: Robust conditional 3d shape generation from casual
  captures.
\newblock 2026.
\newblock URL \url{https://arxiv.org/abs/2601.11514}.

\bibitem[Sani et~al.(2026)Sani, Ku, Jamali, Sani, Khoshtab, Sun, Fazel, Tam,
  Chong, Chan, Tsang, Hsu, Lam, Ng, Chu, Mak, Wu, Wong, Ho, Ruan, Li, Fang,
  Yeh, Cheng, Nie, and Chen]{2603.27862v1}
Samin~Mahdizadeh Sani, Max Ku, Nima Jamali, Matina~Mahdizadeh Sani, Paria
  Khoshtab, Wei-Chieh Sun, Parnian Fazel, Zhi~Rui Tam, Thomas Chong, Edisy
  Kin~Wai Chan, Donald Wai~Tong Tsang, Chiao-Wei Hsu, Ting~Wai Lam, Ho~Yin~Sam
  Ng, Chiafeng Chu, Chak-Wing Mak, Keming Wu, Hiu~Tung Wong, Yik~Chun Ho, Chi
  Ruan, Zhuofeng Li, I-Sheng Fang, Shih-Ying Yeh, Ho~Kei Cheng, Ping Nie, and
  Wenhu Chen.
\newblock {ImagenWorld}: Stress-testing image generation models with
  explainable human evaluation on open-ended real-world tasks.
\newblock 2026.
\newblock URL \url{https://arxiv.org/abs/2603.27862}.

\bibitem[Kumar et~al.(2026)Kumar, Feng, and Tang]{2603.07990v2}
Bhavesh Kumar, Dylan Feng, and Leonard Tang.
\newblock {MJ1: Multimodal Judgment via Grounded Verification}.
\newblock 2026.
\newblock URL \url{https://arxiv.org/abs/2603.07990}.

\bibitem[Zheng et~al.(2023)Zheng, Chiang, Sheng, Zhuang, Wu, Zhuang, Lin, Li,
  Li, Xing, Zhang, Gonzalez, and Stoica]{2306.05685}
Lianmin Zheng, Wei-Lin Chiang, Ying Sheng, Siyuan Zhuang, Zhanghao Wu, Yonghao
  Zhuang, Zi~Lin, Zhuohan Li, Dacheng Li, Eric~P. Xing, Hao Zhang, Joseph~E.
  Gonzalez, and Ion Stoica.
\newblock {Judging LLM-as-a-Judge with MT-Bench and Chatbot Arena}.
\newblock 2023.
\newblock URL \url{https://arxiv.org/abs/2306.05685}.

\bibitem[Wang et~al.(2023)Wang, Li, Chen, Cai, Zhu, Lin, Cao, Liu, Liu, and
  Sui]{2305.17926}
Peiyi Wang, Lei Li, Liang Chen, Zefan Cai, Dawei Zhu, Binghuai Lin, Yunbo Cao,
  Qi~Liu, Tianyu Liu, and Zhifang Sui.
\newblock {Large Language Models are not Fair Evaluators}.
\newblock 2023.
\newblock URL \url{https://arxiv.org/abs/2305.17926}.

\bibitem[Li et~al.(2025)Li, Zheng, Rupprecht, and Vedaldi]{2503.22677v2}
Ruining Li, Chuanxia Zheng, Christian Rupprecht, and Andrea Vedaldi.
\newblock {DSO}: Aligning 3d generators with simulation feedback for physical
  soundness.
\newblock 2025.
\newblock URL \url{https://arxiv.org/abs/2503.22677}.

\bibitem[Liu et~al.(2025)Liu, Liu, Zhang, and Jia]{2506.15684v1}
Qingming Liu, Zhen Liu, Dinghuai Zhang, and Kui Jia.
\newblock {Nabla-R2D3}: Effective and efficient 3d diffusion alignment with 2d
  rewards.
\newblock 2025.
\newblock URL \url{https://arxiv.org/abs/2506.15684}.

\bibitem[Zhou et~al.(2025)Zhou, Xia, Ma, Fan, Yang, and Chua]{2502.04370v1}
Zhenglin Zhou, Xiaobo Xia, Fan Ma, Hehe Fan, Yi~Yang, and Tat-Seng Chua.
\newblock {DreamDPO}: Aligning text-to-3d generation with human preferences via
  direct preference optimization.
\newblock 2025.
\newblock URL \url{https://arxiv.org/abs/2502.04370}.

\bibitem[Chen et~al.(2025)Chen, Wang, Chen, Ye, Shi, Zhao, Zhao, Qu, Lin, Shen,
  Kale, Essa, and Shi]{2511.20629v5}
Chieh-Yun Chen, Zhonghao Wang, Qi~Chen, Zhifan Ye, Min Shi, Yue Zhao, Yinan
  Zhao, Hui Qu, Wei-An Lin, Yiru Shen, Ajinkya Kale, Irfan Essa, and Humphrey
  Shi.
\newblock {MapReduce LoRA}: Advancing the pareto front in multi-preference
  optimization for generative models.
\newblock 2025.
\newblock URL \url{https://arxiv.org/abs/2511.20629}.

\bibitem[Truong et~al.(2025)Truong, Mahmoud, Lukovi\'c, and
  Solomon]{2504.15933v2}
Anh Truong, Ahmed~H. Mahmoud, Mina~Konakovi\'c Lukovi\'c, and Justin Solomon.
\newblock {Low-Rank Adaptation of Neural Fields}.
\newblock 2025.
\newblock URL \url{https://arxiv.org/abs/2504.15933}.

\bibitem[Bai et~al.(2025)Bai, Chen, Liu, Wang, Ge, Song, Dang, Wang, Wang,
  Tang, Zhong, Zhu, Yang, Li, Wan, Wang, Ding, Fu, Xu, Ye, Zhang, Xie, Cheng,
  Zhang, Yang, Xu, and Lin]{2502.13923}
Shuai Bai, Keqin Chen, Xuejing Liu, Jialin Wang, Wenbin Ge, Sibo Song, Kai
  Dang, Peng Wang, Shijie Wang, Jun Tang, Humen Zhong, Yuanzhi Zhu, Mingkun
  Yang, Zhaohai Li, Jianqiang Wan, Pengfei Wang, Wei Ding, Zheren Fu, Yiheng
  Xu, Jiabo Ye, Xi~Zhang, Tianbao Xie, Zesen Cheng, Hang Zhang, Zhibo Yang,
  Haiyang Xu, and Junyang Lin.
\newblock {Qwen2.5-VL Technical Report}.
\newblock arXiv:2502.13923 [cs.CV], 2025.
\newblock URL \url{https://arxiv.org/abs/2502.13923}.

\bibitem[Zhu et~al.(2025)Zhu, Wang, Chen, Liu, Ye, Gu, Tian, Duan, Su, Shao,
  Gao, Cui, Wang, Cao, Liu, Wei, Zhang, Wang, Xu, Li, Wang, Deng, Li, He,
  Jiang, Luo, Wang, He, Shi, Zhang, Shao, He, Xiong, Qu, Sun, Jiao, Lv, Wu,
  Zhang, Deng, Ge, Chen, Wang, Dou, Lu, Zhu, Lu, Lin, Qiao, Dai, and
  Wang]{2504.10479}
Jinguo Zhu, Weiyun Wang, Zhe Chen, Zhaoyang Liu, Shenglong Ye, Lixin Gu, Hao
  Tian, Yuchen Duan, Weijie Su, Jie Shao, Zhangwei Gao, Erfei Cui, Xuehui Wang,
  Yue Cao, Yangzhou Liu, Xingguang Wei, Hongjie Zhang, Haomin Wang, Weiye Xu,
  Hao Li, Jiahao Wang, Nianchen Deng, Songze Li, Yinan He, Tan Jiang, Jiapeng
  Luo, Yi~Wang, Conghui He, Botian Shi, Xingcheng Zhang, Wenqi Shao, Junjun He,
  Yingtong Xiong, Wenwen Qu, Peng Sun, Penglong Jiao, Han Lv, Lijun Wu, Kaipeng
  Zhang, Huipeng Deng, Jiaye Ge, Kai Chen, Limin Wang, Min Dou, Lewei Lu,
  Xizhou Zhu, Tong Lu, Dahua Lin, Yu~Qiao, Jifeng Dai, and Wenhai Wang.
\newblock {InternVL3: Exploring Advanced Training and Test-Time Recipes for
  Open-Source Multimodal Models}.
\newblock arXiv:2504.10479 [cs.CV], 2025.
\newblock URL \url{https://arxiv.org/abs/2504.10479}.

\bibitem[Radford et~al.(2021)Radford, Kim, Hallacy, Ramesh, Goh, Agarwal,
  Sastry, Askell, Mishkin, Clark, Krueger, and Sutskever]{2103.00020}
Alec Radford, Jong~Wook Kim, Chris Hallacy, Aditya Ramesh, Gabriel Goh,
  Sandhini Agarwal, Girish Sastry, Amanda Askell, Pamela Mishkin, Jack Clark,
  Gretchen Krueger, and Ilya Sutskever.
\newblock {Learning Transferable Visual Models From Natural Language
  Supervision}.
\newblock arXiv:2103.00020 [cs.CV], 2021.
\newblock URL \url{https://arxiv.org/abs/2103.00020}.

\bibitem[Ilharco et~al.(2021)Ilharco, Wortsman, Wightman, Gordon, Carlini,
  Taori, Dave, Shankar, Namkoong, Miller, Hajishirzi, Farhadi, and
  Schmidt]{ilharco_openclip}
Gabriel Ilharco, Mitchell Wortsman, Ross Wightman, Cade Gordon, Nicholas
  Carlini, Rohan Taori, Achal Dave, Vaishaal Shankar, Hongseok Namkoong, John
  Miller, Hannaneh Hajishirzi, Ali Farhadi, and Ludwig Schmidt.
\newblock {OpenCLIP}.
\newblock Zenodo, software (version 0.1), 2021.
\newblock URL \url{https://doi.org/10.5281/zenodo.5143773}.

\bibitem[Bradley and Terry(1952)]{bradley1952}
Ralph~Allan Bradley and Milton~E. Terry.
\newblock {Rank Analysis of Incomplete Block Designs: I. The Method of Paired
  Comparisons}.
\newblock \emph{Biometrika}, 39\penalty0 (3/4):\penalty0 324--345, 1952.

\bibitem[Pedregosa et~al.(2011)Pedregosa, Varoquaux, Gramfort, Michel, Thirion,
  Grisel, Blondel, Prettenhofer, Weiss, Dubourg, Vanderplas, Passos,
  Cournapeau, Brucher, Perrot, and Duchesnay]{scikit-learn}
Fabian Pedregosa, Ga{\"e}l Varoquaux, Alexandre Gramfort, Vincent Michel,
  Bertrand Thirion, Olivier Grisel, Mathieu Blondel, Peter Prettenhofer, Ron
  Weiss, Vincent Dubourg, Jake Vanderplas, Alexandre Passos, David Cournapeau,
  Matthieu Brucher, Matthieu Perrot, and {\'E}douard Duchesnay.
\newblock {Scikit-learn: Machine Learning in Python}.
\newblock \emph{Journal of Machine Learning Research}, 12:\penalty0 2825--2830,
  2011.

\bibitem[Downs et~al.(2022)Downs, Francis, Koenig, Kinman, Hickman, Reymann,
  McHugh, and Vanhoucke]{2204.11918}
Laura Downs, Anthony Francis, Nate Koenig, Brandon Kinman, Ryan Hickman, Krista
  Reymann, Thomas~B. McHugh, and Vincent Vanhoucke.
\newblock {Google Scanned Objects: A High-Quality Dataset of 3D Scanned
  Household Items}.
\newblock arXiv:2204.11918 [cs.RO], 2022.
\newblock URL \url{https://arxiv.org/abs/2204.11918}.

\bibitem[Boss et~al.(2024)Boss, Huang, Vasishta, and Jampani]{2408.00653}
Mark Boss, Zixuan Huang, Aaryaman Vasishta, and Varun Jampani.
\newblock {SF3D: Stable Fast 3D Mesh Reconstruction with UV-unwrapping and
  Illumination Disentanglement}.
\newblock arXiv:2408.00653 [cs.CV], 2024.
\newblock URL \url{https://arxiv.org/abs/2408.00653}.

\bibitem[Tochilkin et~al.(2024)Tochilkin, Pankratz, Liu, Huang, Letts, Li,
  Liang, Laforte, Jampani, and Cao]{2403.02151}
Dmitry Tochilkin, David Pankratz, Zexiang Liu, Zixuan Huang, Adam Letts,
  Yangguang Li, Ding Liang, Christian Laforte, Varun Jampani, and Yan-Pei Cao.
\newblock {TripoSR: Fast 3D Object Reconstruction from a Single Image}.
\newblock arXiv:2403.02151 [cs.CV], 2024.
\newblock URL \url{https://arxiv.org/abs/2403.02151}.

\bibitem[Wilson(1927)]{wilson1927}
Edwin~B. Wilson.
\newblock {Probable Inference, the Law of Succession, and Statistical
  Inference}.
\newblock \emph{Journal of the American Statistical Association}, 22\penalty0
  (158):\penalty0 209--212, 1927.

\bibitem[Efron(1979)]{efron1979}
Bradley Efron.
\newblock {Bootstrap Methods: Another Look at the Jackknife}.
\newblock \emph{The Annals of Statistics}, 7\penalty0 (1):\penalty0 1--26,
  1979.

\bibitem[Cohen(1960)]{cohen1960}
Jacob Cohen.
\newblock {A Coefficient of Agreement for Nominal Scales}.
\newblock \emph{Educational and Psychological Measurement}, 20\penalty0
  (1):\penalty0 37--46, 1960.

\bibitem[Landis and Koch(1977)]{landis1977}
J.~Richard Landis and Gary~G. Koch.
\newblock {The Measurement of Observer Agreement for Categorical Data}.
\newblock \emph{Biometrics}, 33\penalty0 (1):\penalty0 159--174, 1977.

\end{thebibliography}

\end{document}